\documentclass[journal]{IEEEtran}
\usepackage{widetext}
\usepackage[normalem]{ulem}

\ifCLASSINFOpdf
  \usepackage[pdftex]{graphicx}
  \graphicspath{{../pdf/}{../jpeg/}}
  \DeclareGraphicsExtensions{.pdf,.jpeg,.png}
\else
  \usepackage[dvips]{graphicx}
  \graphicspath{{../eps/}}
  \DeclareGraphicsExtensions{.eps}
\fi
\usepackage{xcolor}

\usepackage{amsmath}
\usepackage{amsfonts}
\usepackage{bm}

\usepackage{subcaption}

\begin{document}
%
\title{On scale-invariant properties in natural images and their simulations \\
\vspace{0.5cm}
\large{\bf Huawei Technical Report \\}}

%
%
%

\author{Maxim Koroteev and Kirill Aistov
\thanks{Huawei Moscow Research Center, Moscow, Russia. Cloud BU.  E-mail: koroteev.maxim@huawei.com}}
\maketitle

\begin{abstract}
We study samples of natural images for which a set of statistical characteristics is computed and scale-invariant properties of samples are demonstrated computationally. Computations of the power spectrum are carried out and a power-law decaying power spectrum is observed on samples taken from van Hateren images of natural scenes. We propose a dynamic model to reproduce the observed slope in the power spectrum qualitatively. For two types of sources for this model the behaviour of power spectrum is investigated and scale-invarinace confirmed numerically. We then discuss potential applications of scale-invariant properties of natural images.
\end{abstract}


%
\IEEEpeerreviewmaketitle

\section{Introduction}
%
%
%
%

\IEEEPARstart{I}{n} recent years with the advent of AI era our possibilities to work with images increased tremendously. Methods of machine learning find their place in multiple applications related to image and video processing.

Nevertheless when talking about some fundamental properties of images and how the image information is processed in the brain the methods of AI are of much moderate help. The reasons for that are clear: generation of a huge number of features can help us {\it ad hoc} but does not allow to expand our understanding of more essential properties of images and their structure.

Putting aside the longlasting controversy of ML advocates and sceptics we just note that the general direction of this work poses us rather closer to the latter than to the former. We are intereseted in studying basic properties of images and would like to invest our efforts to apply these fundamental properties to practical tasks such as image quality assessment, image and video compression etc. This task sounds as too generic but in this paper we make just several first steps in this direction and are not aiming at contributing much to the novelty of the results; in several parts we found it relevant to reproduce results obtained some twenty years ago but attracting much wider material and trying to further analyze them.

The main motivation to carry out these simulations is realization of the fact that fundamental properties of natural images discovered long ago are still not sufficiently utilized in practical applications; the search of new approaches to this problem by means of modeling is also one of the main goals of this work.


\section{Preprocessing of the Data sets}
\label{sec:datasets}

We exploited several image data sets widely spread in the computer science community. Some more details about each data set would be added later but here we stress some generic information. It seems reasonable to mention from the very beginning that what we were trying to avoid was collecting as many images as possible and whatever possible, the approach widely propagated in modern research and, we can admit, quite natural in the sense of AI applications. But from the point of view of pure experiment and an effort to understand the relations between the scales of objects on images and statistical functions it seems more reasonable to study images sets with significant fraction of uniformity, i.e., the scenes should show sufficient amount of similar objects.

Among such data sets we found one which appears to be the most suitable and that was generated some twenty years ago by van Hateren\cite{vanHateren}. It represents a series of still images portreying natural scenes. This data set provides a significant degree of uniformity in the data as it does not contain any artificial objects. On the other hand, multiple natural objects captured on these images represent a range of various sizes, e.g., trees, streams, stones, bushes, leaves etc. This dataset is usually provided in two forms: linearized and calibrated; the details of the calibration procedure are presented in \cite{vanHateren}. 

For our computations we used not the calibrated luminance of the images but made a trasnform to the image contrast, which we define in a particular point $x$ of an image as \cite{RudermanBialek, RudermanBialek2}
$$
\phi(x) = \ln\left(I / I_{0}\right),
$$
where $I(x)$ is the intensity in a given point and normalization factor $I_{0}$ is chosen for each image so as $\sum\phi(x) =0$ over the image.

\section{Computational Results for Image Domain}
\label{sec:image_domain_results}

\begin{figure}
\centering
\includegraphics[width=1.01\linewidth]{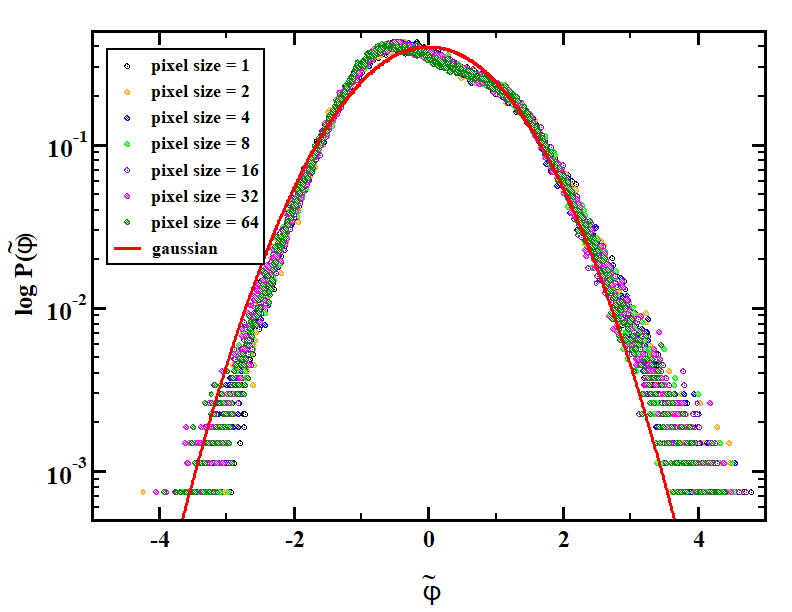}
\caption{Pixel-value distributions for linearized van Hateren data set ($\approx 4000$ images)\cite{vanHateren} for various pixel size. From each image $32$ patches of size $N \times N$ were sampled. Pixel size = N means that each patch of the size $N \times N$ was replaced by its average and then the distribution of such pixels was computed. Thus the pixels on various length scales were constructed. The sample was standartized as described in the main text so that $\mu_{\tilde{\phi}}=0$, $\sigma_{\tilde{\phi}}=1$ and all distributions were normalized to sum up to the unit. Note the log scale on the y-axis. Gaussian distribution $N(0,1)$ is provided for comparison (red). The results for calibrated van Hateren images do not show any noticeable difference.}
\label{fig:pixel_distrib}
\end{figure}

Following \cite{Ruderman} we started by measuring pixel distributions in a sample of van Hateren images. Specifically we take an image in the data set and sample a set of patches of size $N\times N$ from this image; this procedure was repeated for {\it each image in the data set}. For each patch we then computed the average of the contrast $\phi$ and this average represents a pixel value {\it on the length scale $N$}. Thus we effectively collected a sample of random pixels for each scale $N$ and were able to compute the distribution of pixels on different length scales. To plot the distributions the sample was standartized by the standard deviation $\sigma_{\phi}$ so that $\tilde{\phi}=\phi / \sigma_{\phi}$; these distributions are shown in fig. \ref{fig:pixel_distrib}. As it was noted in \cite{RudermanBialek, Ruderman} and as we were able to confirm on fig. 1 the pixel distribution scales wrt. the pixel size. Then, as we plot the distributions in the semi-log scale we notice that the tails of the distribution tend to be straight, especially noticeable in the right part of the tail (\ref{fig:pixel_distrib}), which indicates the proximity of the distributions to exponential ones while if the pixels had been independent, the central limit theorem would have suggested gaussian distributions for pixels, which would have demonstrated much sharper decline than we observe. 
This observation is not surprising; it was construed by the authors\cite{Ruderman, RudermanBialek2} as an evidence of correlations existing between adjacent pixels (see also \cite{Gerhard}). More importantly, of course, is that these correlations are observed on various scales and turn out to be invariant wrt. scales as it is seen from the distributions in fig. \ref{fig:pixel_distrib}.

\begin{figure}
\centering
\includegraphics[width=1.01\linewidth]{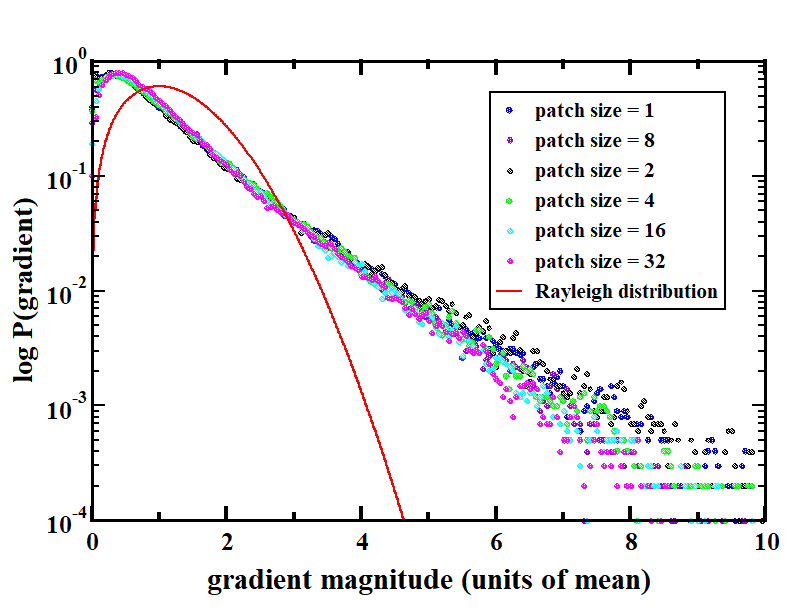}
\caption{Gradient distributions for linearized van Hateren data set ($\approx 4000$ images ) \cite{vanHateren} for various pixel size. The pixels are computed as explained in the legend for fig. \ref{fig:pixel_distrib}. The gradient computation is done using central differences as explained in the main text. All distributions were re-scaled to sum up to $1$. Rayleigh distribution $x \exp\{-x^{2}/2\}$, where $x$ is the magnitude of the gradient is provided for comparison. Note the semi-log scale to demonstrate longer tails for the distributions.}
\label{fig:gradient_distrib}
\end{figure}

In a similar way we carried out the computation of pixel gradients on various scales. We randomly sampled a patch of the size $N\times N$ in the image and computed the pixel values for its right, left, top, and bottom neighbors as we explained above. Then we computed two derivatives at this point simply as a pixel half central difference in the corresponding directions. After that the absolute value of the gradient is computed using these two derivatives. The results of this computation are shown in fig. \ref{fig:gradient_distrib}.
For comparison we provide the Rayleigh distribution, which we could have expected for the gradient if both components had been distributed normally, independently and with the same variance (see also \cite{RudermanBialek2}). The longer tails in the gradient distributions compared to those which one would have expected for the gaussian distributed pixels indicate higher proportion of big differences between pixels which may be explained by image edges.
 
Both computations demonstrate scale-invariant behaviour and longer tails compared to expected for gaussian distributed variables. 

\section{Computational results for Fourier Domain}
\label{sec:fourier_domain_results}

\begin{figure}
\centering
\includegraphics[width=1.0\linewidth]{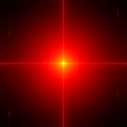}
\caption{Log of the 2D power spectrum computed over a sample of randomly chosen $128\times 128$ pixels patches for van Hateren images. The patches are collected in the same way as explained in the legends for figs. \ref{fig:pixel_distrib} and \ref{fig:gradient_distrib}. The spectrum was computed over each patch and then the average was computed in each point. Then the spectrum was shifted so that zero frequency would correspond to the center of the patch. The horizontal and vertical central lines are accounted by the edges of images and should be neglected.}
\label{fig:2d_spectrum}
\end{figure}

To look further into the scaling properties of the images we computed Fourier power spectra. The patches were collected from random places of an image sample in the same way as before and the power spectrum was computed for each patch. After that these 2D spectra were averaged over the sample; the results for the averaged 2D spectrum are presented in fig. \ref{fig:2d_spectrum}.

\begin{figure}
\centering
\includegraphics[width=0.95\linewidth]{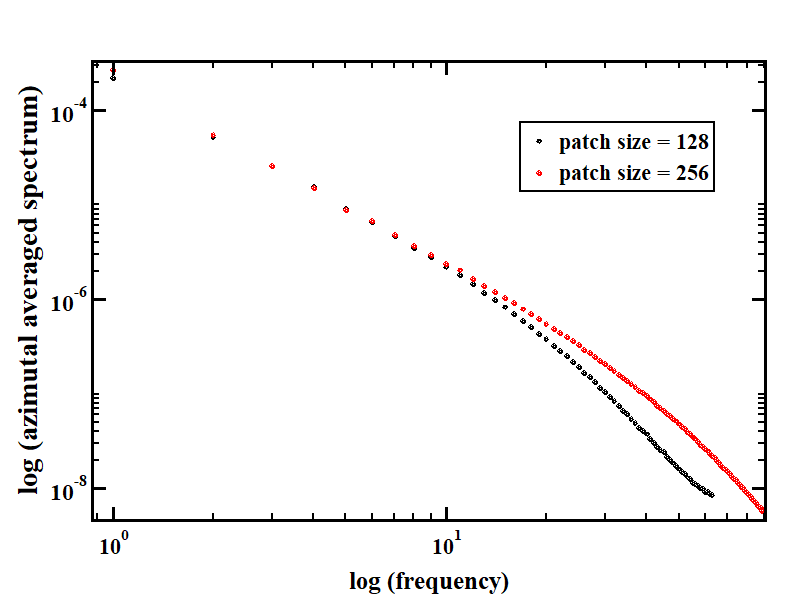}
\caption{1D power spectra for a sample of patches collected from van Hateren image set\cite{vanHateren}; note double log scale. 12000 patches were collected and 2D spectrum was computed for each of them (an example of such 2D spectrum is provided in fig. \ref{fig:2d_spectrum}). 2D spectra were averaged at each point and then additionally an azimuthal average was computed to obtain 1D power spectrum shown in this figure. The spectrum demonstrates power-law scaling on the more than the order of magnitude over frequency and four orders of magnitudes over the spectrum. The slope in the power-law regime is $\approx -2$.}
\label{fig:1d_spectrum}
\end{figure}

Using this computation we then were able to compute 1D power spectrum by azimuthal averaging the 2D spectrum; the results are given in fig. \ref{fig:1d_spectrum} and demonstrate power-law regime in the power spectrum spanning at least an order of magnitude over frequencies.
The slope of the power spectrum was close to $-2$ and corresponds to the observations made earlier by multiple authors during several decades. This power-law behaviour is in correspondence with the distributions shown in figs. \ref{fig:pixel_distrib}, \ref{fig:gradient_distrib} in the sense that the deviation those from gaussian or Raleigh (for gradients) is a sign of correlations presented in the natural images and reflected by the power-law power spectrum.

\section{A model for scale-invariant power spectrum of natural images}

Looking at the specific behavior of the statistics for natural images it is reasonable to ask whether we can catch these properties applying a stochastic model. Some models of that kind were previously proposed in \cite{RudermanOrigin, Mumford} and we in this paper try to further analyze those models by varying the type of the source and focusing on spectral properties of synthetic images. One of the key ideas of the approach related to these models is that if we observe a power-law spectrum for patches of natural images the power-law behavior should be observed in correlation functions. Such power-law correlations may be a result of a natural structure of images, as it seems plausible to conjecture that power-law spectra observed for natural images occur as a result of intersections of multiple 2D objects resulting in emerging multiple edges of various lengths.

\begin{figure}
\centering
\includegraphics[width=0.95\linewidth]{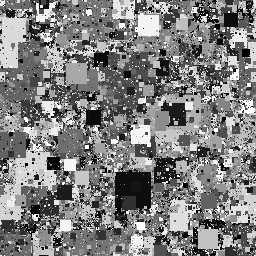}
\caption{Random patches of the size $N\times N$, $p(N)\sim 1/N^{\gamma}$, $\gamma=3$ were generated and uniformly distributed in the image of the size $256\times 256$ pixels. The patch sizes are between $1$ and $N/2$. The color of patches is chosen randomly using a uniform distribution in the interval $[0, 255]$.}
\label{fig:synthetic_image}
\end{figure}

Having this in mind and motivated by consideration of power-law fragmentation sources for 1D sequences (see e.g.,  \cite{KoroteevMiller2011}) we consider the following, sufficiently elementary model (for more general treatment see \cite{Mumford}). Let us assume for the sake of simplicity we have a square image of some size and choose some contrast range $[0, I_{max}]$; we can think of it as a range of integer numbers. We generate a random patch of the size $N\times N$, where N is taken from a power-law distribution $p(N)\sim 1/N^{\gamma}$ and insert it into the randomly chosen position of the image applying periodic boundary conditions. The contrast of the patch is also picked up randomly from the uniform distribution on the interval $[0, I_{max}]$. This process continues sufficiently long time to achieve a stationary state. This procedure results in generating synthetic images, one of them is shown in fig. \ref{fig:synthetic_image}.   

\begin{figure}
\centering
\includegraphics[width=0.98\linewidth]{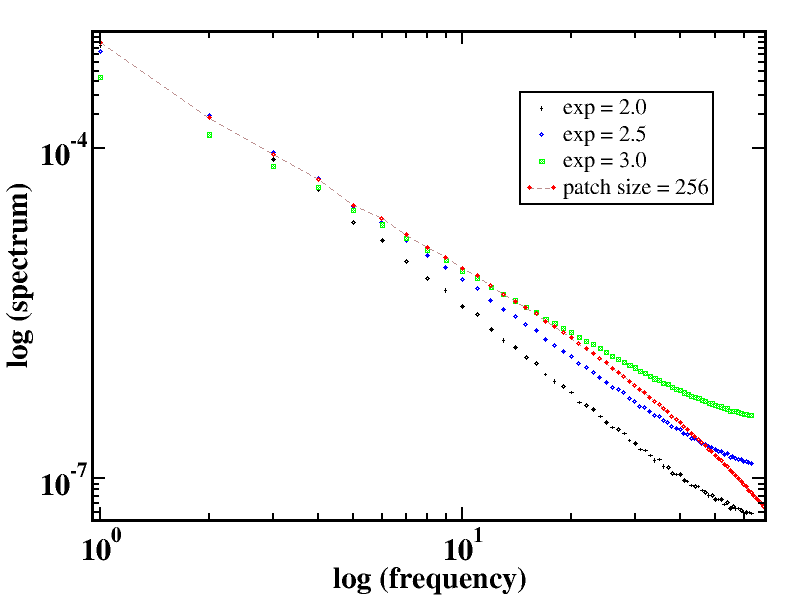}
\caption{Power spectra for a sample of synthetic images generated using random power-law sources of the form $p\sim 1/N^{\gamma}$ with various exponents $\gamma$. The sample consisted of $100$ images of the size $512\times 512$ pixels. Random patches were sampled in a standard way explained above in the main text. For comparison a power spectrum for a natural sample of images is presented (red points); note, this spectrum was shifted up by a factor of $\approx 3.5$. }
\label{fig:synthetic_spectrum}
\end{figure}

We then were looking at the power spectrum for a set of synthetic images in fig. \ref{fig:synthetic_image} in the same way as we did for natural images; the comparison of previosuly computed power spectrum for van Hateren images with the synthetic one is shown in fig. \ref{fig:synthetic_spectrum}.
It is seen that depending on the exponent of the power-law source the slope of the resulting power spectrum varies. Larger exponents (sharper slopes of the source) imply smaller mean values for generated patches. From the point of view of the spectrum this makes it more decorrelated: it becomes less probable to find pixels of the same color close to each other. This evidently results in a flatter spectrum. If we decrease the exponent $\gamma$ the spectrum drops sharper and approaches to a singular one, with one frequency.

\section{Analysis of the model}

The model with a power-law source described above has a property of being able to generate a power-law power spectrum and also is able to reproduce qualitatively the power spectra observed in real images by varying the exponent in the source as it is seen in fig. \ref{fig:synthetic_spectrum}. The reason for emerging a scale-invariant power spectrum both in natural and synethetic images was the subject of much research and debate in previous years \cite{Olshausen}. Among several explanations one suggested that the observed behavior of the power spectrum may be related to the statistic of edges of various sizes in the synthetic images produced by the model. We investigae this question in more detail.

\begin{figure}
\centering
\includegraphics[width=0.99\linewidth]{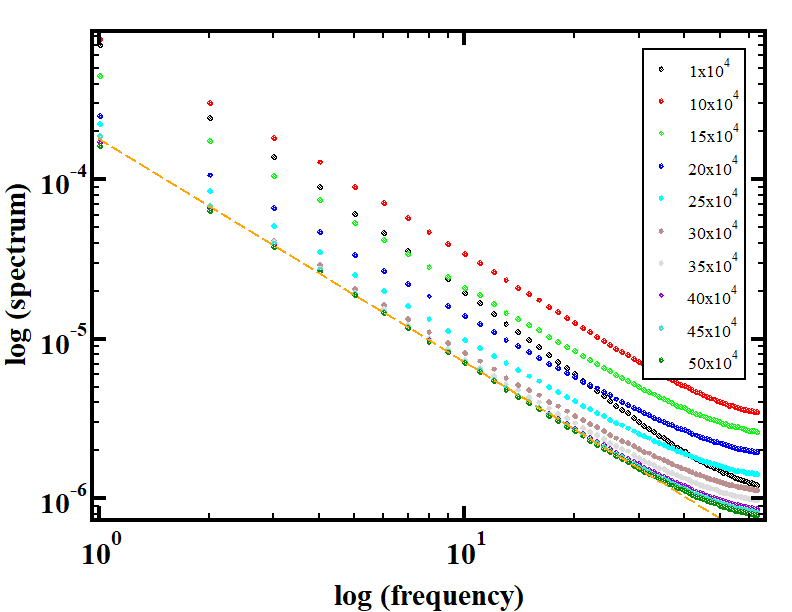}
\caption{Converegnce of power spectra to the stationary state when increasing the number of generated patches or equivalently when increasing the time (explained in the main text). $100$ images of the size $512\times 512$ pixels were generated with our model applying the power-law source with $\gamma=3$. For each image the process of generating new patches continued the number of times indicated in the legend box. Then $30$ patches of the size $128\times 128$ pixels were collected from each of $100$ images and the usual procedure for spectrum computation was applied. The final slope is close to $-2$ }
\label{fig:spectrum_convergence}
\end{figure}

To begin we first make a note about time scales in the model under consideration. The number of times we repeat the process of choosing a random patch using our source distribution results in different power spectra. If we assume that one step of the process of generating a random size patch with randomly chosen color corresponds to a unit of time, the observed variation of the spectrum implies the system under consideration is non-stationary and the natural question to ask is whether we are able to attain a stationary state in terms of the spectrum stabilization. The result of this computation is shown in fig. \ref{fig:spectrum_convergence}.
It clearly shows the spectra for small times demonstrate different slopes which change after the appropriate number of time steps; after that the amplitude of the spectrum also stabilizes. It seems reasonable to assume a certain dependence of the stabilization time for the spectrum on the size of the image and the power of the source.

\begin{figure}
\centering
\includegraphics[width=0.95\linewidth]{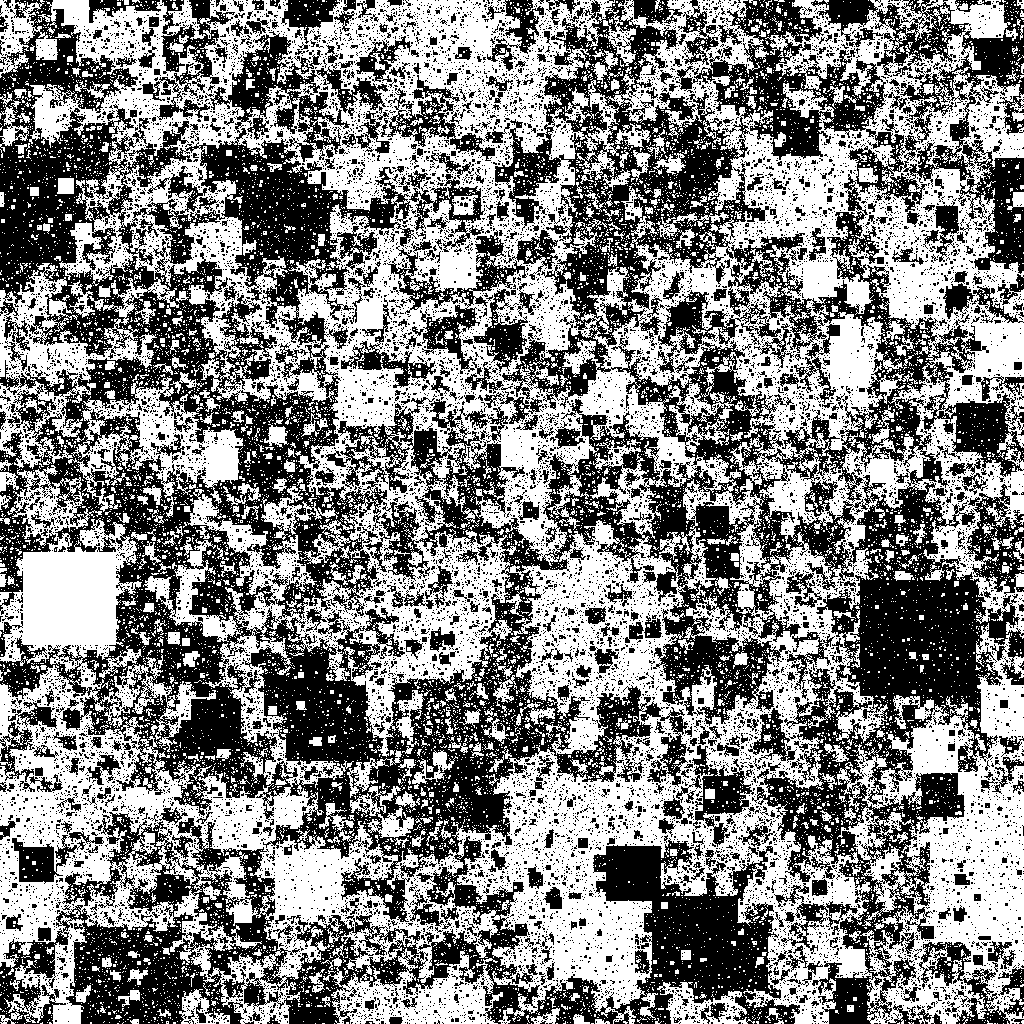}
\caption{An example of a synthetic binary image generated using the model proposed in the main text. The exponent of the source is $\gamma=3.3$, the size of the image $1024\times 1024$ pixels. The model was simulated until the stationary state for the spectrum was achieved; the image shows the state of the model in this stationary regime.}
\label{fig:binary_image}
\end{figure}

\begin{figure}
\centering
\includegraphics[width=1.0\linewidth]{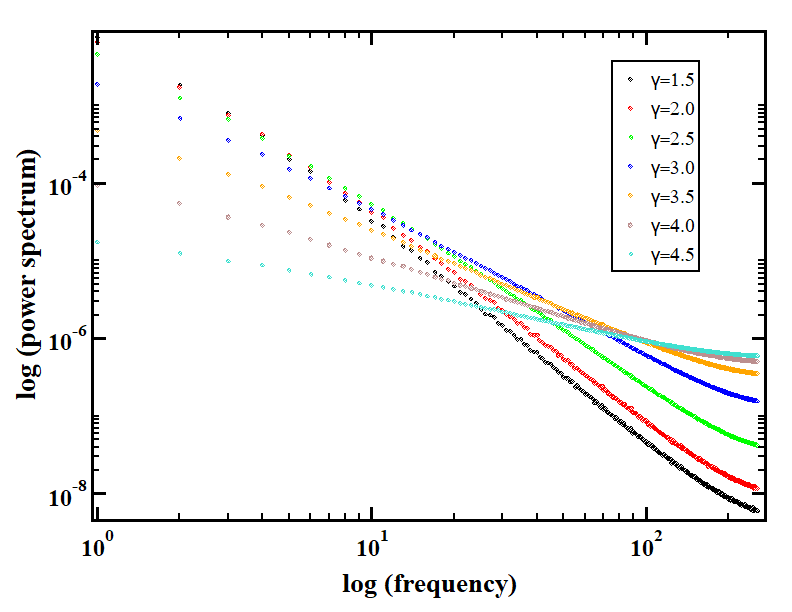}
\caption{Behavior of the power spectrum for binary images depending on the exponent of the source. The figure shows stationary-state power spectra obtained for the model with different exponents of the source. The image size was $1024\times1024$ pixels. Note the size of patches generated by the source is restricted by $N/2\times N/2$, where $N\times N$ is the system size, so $N/2=512$ in this case.}
\label{fig:binary_spectra}
\end{figure}

Keeping this in mind, we will focus on stationary states of the model in terms of the power spectrum which always can be obtained empirically after the sufficient number of time steps and consider a simplification of the model to investigate our assumption above, that the behavior of the power spectrum is determined by a certain statistic of edges in the image. It is clear that the uniform distribution  of colors over a significant range of numbers, e.g., $[0, 255]$ may not be essential for forming edges, the {\it difference} of colors being a more significant property. Therefore we can consider the same model as before but instead of using $256$ colors for patches we now take only two colors, black and white, i.e., we consider binary images; an example of such synthetic image is given in fig. \ref{fig:binary_image}. 
For binary images we also observe convergence to the stationary state in terms of the power spectrum. Note that while fig. \ref{fig:spectrum_convergence} demonstrates the convergence of the spectrum to the slope $\approx -2$ for the source exponent $\gamma=3$, now we make computations for a range of $\gamma$ to investigate the dependence. We obtain various slopes for the stationary states depending on the exponent of the patch source (fig. \ref{fig:binary_spectra}).
It is seen that larger exponents of the source make the spectrum flatter, which is accounted for by generating on average smaller patches with the source in this case; this results in similar magnitudes of spectral components for various frequences making the spectrum noise-like but keeping the scale invariant structure. On the other hand, smaller exponents in the sourse imply that large size patches become more probable. This occurs, as fig. \ref{fig:binary_spectra} shows, for $\gamma=1.5$ and $2.0$ which accounts for the similar behavior. It seems plausible to assume that the system size may affect the behavior in this case; the more detailed consideration requires to increase the size of the system.

From the characteristic scales point of view the increase of value of the source exponent $\gamma$ decreases the average patch size; for a fixed system size this can be thought of as zooming out the scene in the image. Similarly, the increasing gamma effectively provides zooming in for the image. Overall we can indicate four characteristic length scales in the problem: $L$ - the system size or the size of the image; $l$ - the size for the spectrum computing area or the patch size; average patch size $M$ generated by the source; pixel size $s$. Presumably it is the relationship between these parameters that determines the observed spectra.

\begin{figure}
\centering
\includegraphics[width=0.99\linewidth]{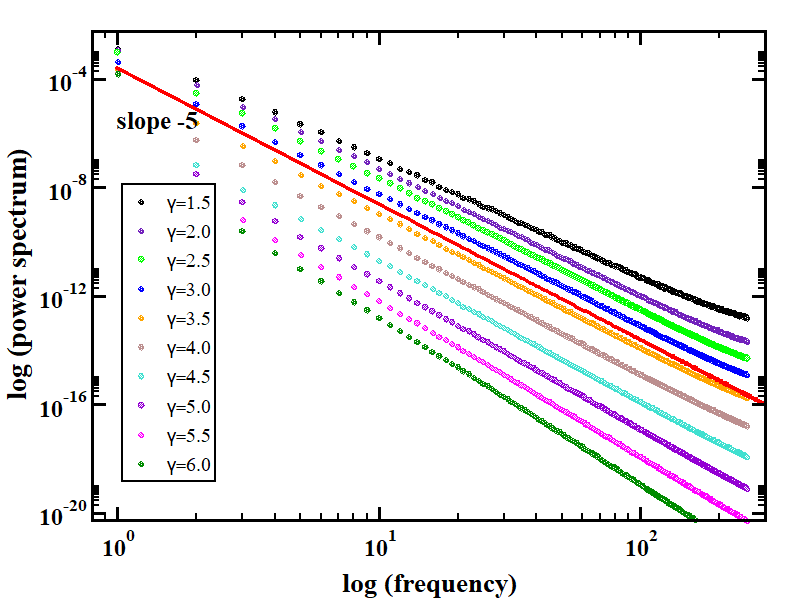}
\caption{Rescaled power spectra from fig. \ref{fig:binary_spectra}. Each spectrum at this figure was obtained by dividing each spectrum from fig. \ref{fig:binary_spectra} by $f^{\gamma}$. For comparison a straight line with the slope $-5$ is shown. Other parameters are the same as in the previous figure.}
\label{fig:binary_spectra_rescaled}
\end{figure}

Another interesting property to note here is the observed scale invariant behavior of the spectrum and the collapse of different slopes. It can be seen, e.g., from fig. \ref{fig:binary_spectra} if we rescale all the spectra as $S(f, L, l, M, s)/f^{\gamma}$, where $f$ is 1D frequency on the x-axis of the figure; the result of this procedure is shown in fig. \ref{fig:binary_spectra_rescaled}.
The slopes of the spectra on the fig. \ref{fig:binary_spectra_rescaled} look similar at least for sufficiently small ratio of the average patch size to the system size and close to $-5$. This allows to assume the following form of the power spectra on fig. \ref{fig:binary_spectra}:
$$
S(f, L, l, M, s) \sim \frac{A(L,l,M, s)}{f^{5-\gamma}}.
$$
The similar behavior is observed when $\gamma$ approaches $5$ but it is worth noting that there may exist a weak dependence of the term $A$ on $f$ which may become dominant when the denominator above approaches the unit. We should stress that the approximation above should be valid only in some regime defined by ratios of parameters in $A$; there may exist other terms not presented above which become leading when $\gamma\to 5$.

In the previous discussion we were focusing on a power-law source. It seems instructive to look to what extent the observations made above remain valid for some other types of the source.

\begin{figure}
\centering
\includegraphics[width=0.95\linewidth]{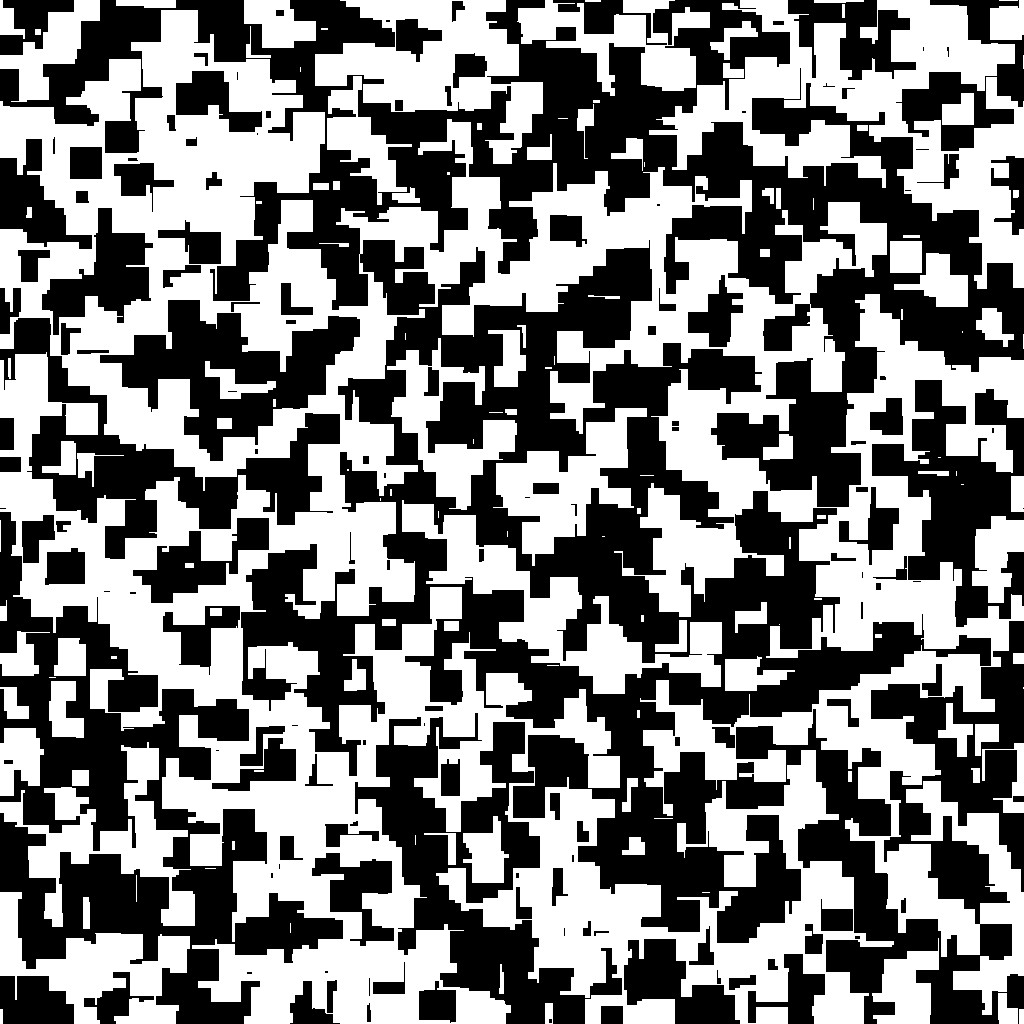}
\caption{An example of a synthetic binary image generated using the monoscale model discussed in the main text. The source generates patches of the fixed size $32\times 32$ pixels; the size of the image $1024\times 1024$ pixels. The model was simulated until the stationary state for the spectrum was achieved; the image shows the state of the model in this stationary regime.}
\label{fig:delta_source}
\end{figure}

\begin{figure}
\centering
\includegraphics[width=0.95\linewidth]{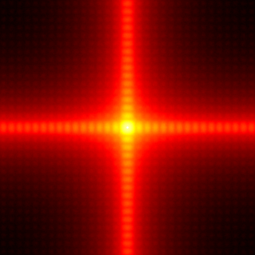}
\caption{2D spectrum for the model with a monoscale source, (fig. \ref{fig:delta_source}). The spectra were computed over patches of the size $256\times 256$. The periodic structure of the spectrum corresponds to the patch size; the nearest peaks in the spectrum occur with the period $8$ over the frequency}
\label{fig:delta_source_spectrum}
\end{figure}

The first candidate for this consideration is a monoscale or $\delta$-source. It implies we generate patches of a fixed size and try to attain a stationary power spectrum, which presumably may be expected after sufficient number of steps. The model of this kind results in synthetic images as one presented in fig. \ref{fig:delta_source}.
The corresponding 2D power spectrum is shown in fig. \ref{fig:delta_source_spectrum}; it demonstrates a distinct periodic structure. This periodicity is in agreement with the length scales observed in fig. \ref{fig:delta_source}. 

\begin{figure}
\centering
\includegraphics[width=0.99\linewidth]{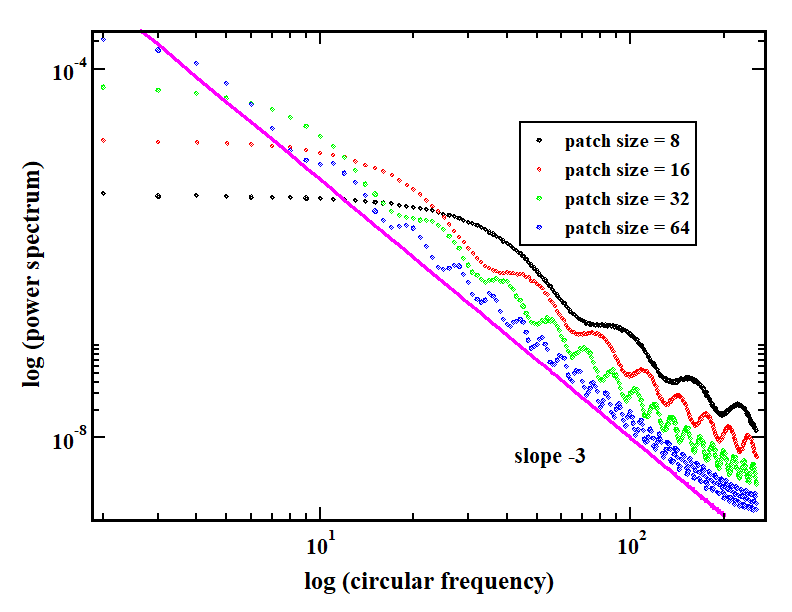}
\caption{1D spectra for the model with a monoscale source for various source patch sizes, (fig. \ref{fig:delta_source}). The spectra were computed over patches of the size $512\times 512$. The periodic structure of the spectrum corresponds to the patch sizes. Slope $-3$ is shown for comparison.}
\label{fig:delta_source_1d_spectrum}
\end{figure}

After averaging the power spectra we obtain 1D spectra as shown in fig. \ref{fig:delta_source_1d_spectrum}.

Interestingly, the resulting slopes for various source patch sizes are close to each other and to the slope $-3$. The part of the specrum on lower frequencies becomes flatter when decreasing the source patch size: quite natural as this corresponds to the synthetic images approaching pixels scattered randomly over the the image.

\section{Discussion}
\label{sec:discussion}

We computationally investigated several properties of natural images which distinctly demonstrate scale invariant behaviour as well as some spectrum related aspects of a dynamic model to reporduce qualitatively the observed power spectra. Note that examples for the models of this sort earlier appeared in some papers (e.g. \cite{RudermanOrigin, Mumford}); we extended this analysis by studying delta sources. It was demonstrated that the power spectra of the model attain a statinary state in which they take on a power-law form and the spectra for natural images can be reproduced at least qualitatively by the stationary state spectra for the synthetic images as the power spectra of those images depend on the $\gamma$ in a sufficiently straightforward way. The question of quantitative agreement of the model with the observed power spectra will be presented elsewhere. In the same time it should be admitted that the formula suggested in the previous section for the synthetic power spectrum can be thought of as a putative one and some related questions which can result to a better approximationg still should be investigated.

In a spatial domain our results for natural images reproduce the observations made earlier in a set of deep and thoughtful papers \cite{RudermanBialek, Ruderman, RudermanOrigin}. In terms of the data we considered much more extensive data sets and were able to confirm scale-invarinace on larger scales. We believe that the material presented in the paper can be used for building more thorough investigations both in terms of computations and models.

When considering a model for power-law power spectra we were motivated, to some extent, by the fragmentation model proposed in \cite{KoroteevMiller2011} in which also several type of sources were proposed for explaining power-law behavior of length distributions in natural genomic sequences. But unlike that model where duplication or fragmentation dynamics, being applied in the framework of natural DNA sequences, allowed obvious physical interpretation, the current model, though being built as a kind of a fragmentation process, does not provide such an interpretation for natural images. It seems obvious that the images of nature we see hardly could be generated with any duplication process occuring in nature. Thus the model has to be thought of as an artificial one. On the other hand, every natural image can be percieved by a human eye and consequently such a model potentially can tell us (if not already told) something about human visual system responsible for transferring image information to the brain.

We were able to propose some simplifications of the model to stress the role of the edges in forming the power-law spectrum and find numeric parameter of the model which allows to control the output slope of the spectrum. Also the detection of stationary state for the spectrum enabled us to focus on the stationary regime but one improtant question left without closer look is the behavior of the model in other regimes far from stationary. This analysis will be the subject of future research.

The analysis presented above raises some questions when looking from the practical point of view. The existence of scale-invariance in samples of natural images poses the question whether these properties are stable to distortions very often occurring in natural images. This could be utilized in many ways, in particular, for the image quality assesment purposes. This problem briefly can be formulated as follows. When looking at an image the human eye (and brain) processes information so to estimate some images as of higher quality than the others. So the question arises if the statistical scale invariant properties of images or functions thereof can be markers of degrading quality of image. 

Nowadays the search of such stable properties or, using the modern terminology of machine learning, features, is mainly carried our using AI methods. The problem of features found in this way is notoriously well-known: they are hard to interpret, they are too many which prevents from developing an intuition about any fundamental property of image quality. Thus the observations presented here as well as in earlier papers indicate potential directions for further research. These directions are under investigation now.


%



\section*{Acknowledgment}

The authors are grateful to their colleagues in Algorithm Innovation Lab for discussions.
\ifCLASSOPTIONcaptionsoff
  \newpage
\fi



\bibliographystyle{IEEEtran}
\bibliography{bibtex/bib/IEEEexample}
%




%




\end{document}